\documentclass{article}
\usepackage[T1]{fontenc} 
\usepackage[utf8]{inputenc} 
\usepackage{ismir,amsmath,cite,url}
\usepackage{graphicx}
\usepackage{color}

\usepackage{lineno}

\usepackage{amsfonts}
\usepackage[rightcaption]{sidecap} 
\sidecaptionvpos{figure}{t}

\usepackage{booktabs}

\usepackage{makecell}

\usepackage{float} 

\usepackage{wrapfig} 

\usepackage{comment} 

\title{Checklist Models for Improved Output Fluency \\ in Piano Fingering Prediction}

\twoauthors
  {Nikita Srivatsan} {Carnegie Mellon University \\ {\tt nsrivats@cmu.edu}}
  {Taylor Berg-Kirkpatrick} {UC San Diego \\ {\tt tberg@eng.ucsd.edu}}

\def\authorname{N. Srivatsan and T. Berg-Kirkpatrick}

\begin{document}

\maketitle
\begin{abstract}
In this work we present a new approach for the task of predicting fingerings for piano music.
While prior neural approaches have often treated this as a sequence tagging problem with independent predictions, we put forward a checklist system, trained via reinforcement learning, that maintains a representation of recent predictions in addition to a hidden state, allowing it to learn soft constraints on output structure.
We also demonstrate that by modifying input representations --- which in prior work using neural models have often taken the form of one-hot encodings over individual keys on the piano --- to encode \textit{relative} position on the keyboard to the prior note instead, we can achieve much better performance.
Additionally, we reassess the use of raw per-note labeling precision as an evaluation metric, noting that it does not adequately measure the \textit{fluency}, i.e. human playability, of a model's output.
To this end, we compare methods across several statistics which track the frequency of adjacent finger predictions that while independently reasonable would be physically challenging to perform in sequence, and implement a reinforcement learning strategy to minimize these as part of our training loss.
Finally through human expert evaluation, we demonstrate significant gains in performability directly attributable to improvements with respect to these metrics.
\end{abstract}
\section{Introduction}

\begin{figure}
    \centering
    \includegraphics[trim= 120 100 160 100, clip, width=0.78\linewidth]{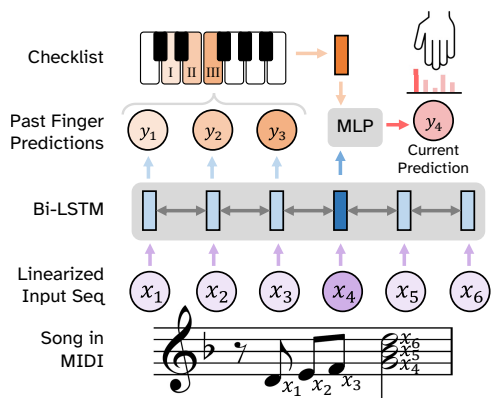}
    \vspace{-1ex}
    \caption{Visualization of the checklist model. Notes are embedded and then passed to a Bi-LSTM which outputs contextualized embeddings at each timestep. A checklist encodes where the fingers that have recently been used are located based on prior predictions. These are both fed into an MLP which predicts the next finger.}
    \label{fig:model}
    \vspace{-2ex}
\end{figure}

While sheet music is often very specific as to \textit{what} notes a musician must play, it can also be vague about \textit{how} to play them.
Composers generally write notation that focuses on describing the desired musical output, but leave the specifics of how to achieve this with their instrument up to the performer, as these details are outside the scope of their role in the creative process or in some cases beyond their expertise.
The piano is an instrument which requires an extensive amount of decision-making on the performer's part.
In order to perform a piece of music, a pianist must either consciously or intuitively select which finger to use to play every note in the song.
Notes may overlap in duration or even have simultaneous onsets.
Since any key on the piano can potentially be played by any finger, there are an exponential number of ways one could perform any given piece; of these however, the majority would be uncomfortable, difficult to play at full tempo, or even physically impossible~\cite{ergonomic}.
The challenge of deciding the most ergonomic fingering to use can be nontrivial for less experienced pianists, who would not yet have the ability to quickly map common musical patterns to conventional fingering strategies, and might produce more consistent and accurate performances if provided with them~\cite{SightReading}.

However recent work has shown that the use of machine learning techniques may be able to assist in this regard~\cite{PIG,goldberg,pitchdiff,BarbaraEXPECTEDRR}.
Automatic piano fingering prediction is the task of inferring finger assignments for each note in a symbolic representation of a piano song, given knowledge of which hand is meant to play each note.
Human players generally prefer fingerings that balance physical comfort and efficiency~\cite{SightReading}, yet these criteria are difficult to quantify and highly subjective~\cite{interview}.
Decisions are simultaneously constrained by the current placement of the musician's fingers, and by how that decision may in turn affect possible options for the notes that follow.
And while the spatial location of the notes on the keyboard is arguably the most salient factor, timing is also crucial; certain fingerings might be easy for a set of consecutive notes, but impossible if the same notes are in a chord.

Prior approaches to this task have largely treated it as a sequence tagging problem, where the input at each step is simply the pitch of the note and the output is a softmax over indices corresponding to each finger, and have employed methods analogous to those used for part of speech tagging such as LSTMs and HMMs.
However, these techniques are limited in their capacity to model output dependencies, and also disregard the amount of elapsed time between notes which can greatly affect a prediction's performability.

We propose an autoregressive approach, where prior predictions are fed back into the network in order to encourage the model to produce fingerings that are not just accurate on average, but also locally \textit{fluent} and therefore playable.
We do this using a checklist which indicates which fingers have either recently been or are currently in use, and where they are on the keyboard relative to the note at the current timestep.
By directly exposing this information, the model can more easily learn to restrict its predictions to fingers that are actually free, and also make decisions more directly influenced by the hand's physical placement.
We also experiment with an approach that only feeds back in the most recent prediction, as well as an autoregressive tagger which maintains a neural representation of prior predictions using an encoder network.

Furthermore, the evaluation metrics used in prior work do not always correlate with what a pianist might intuitively perceive as ``good'' predictions.
In fact there are many types of desirable (or undesirable) patterns that may emerge in a model's output, which a simplistic metric such as labeling precision may not actually reflect.
We demonstrate that it is possible for two models with nearly identical labeling precision to differ dramatically in the frequency of specific types of output subsequences that are physically challenging to play.
Therefore, we evaluate on a battery of metrics that collectively convey a more holistic and interpretable overview of fingering quality.

Since these metrics are not always correlated with the labeling precision that a cross entropy loss optimizes, we also investigate incorporating them explicitly into our loss function at train time.
While these scores are non-differentiable, we show that reinforcement learning techniques---which have previously seen success in sequence generation tasks outside of music---can successfully optimize them.
This, in conjunction with our checklist approach, ensures that the model not only has direct access to the information it would need to avoid undesirable output patterns, but is also encouraged to do so.

Finally, we conduct human evaluations and qualitative analysis which confirm that our approach improves predicted fingerings in ways consistent with our modeling choices, and also suggest directions for future work.

In summary our contributions are as follows:
(1) We provide a comparison of various input representations for LSTM models
(2) We put forward checklist based approaches which directly incorporate information from previous decisions
(3) We introduce several additional metrics which track fluency of output, and demonstrate how to train directly on them using reinforcement learning.

\vspace{-1ex}
\section{Related Work}

While constraint-based models for automatic piano fingering have long existed~\cite{ergonomic}, the standardization as a machine learning task which we follow was formalized by~\cite{PIG}, which introduced the PIano fingerinG (PIG) dataset, put forward LSTM and HMM baselines (following prior work using HMMs~\cite{MergedHMM,AutomaticDecision}), and has since been followed up on by others.
\cite{goldberg} demonstrated the value of pretraining on even a noisy automatically annotated dataset, although they focused on basic LSTM models and evaluated exclusively on labeling precision.
\cite{pitchdiff} put forward the idea of representing inputs based on relative difference in pitch rather than absolute pitch, and also showed improvements from the use of a constrained transition matrix, albeit one that was built off of prior knowledge of the task.
\cite{BarbaraEXPECTEDRR} recast fingering as an information retrieval problem, although they focused on the Czerny corpus instead of the PIG dataset.

Our work also draws from research into implicitly learning output structure by feeding prior decisions into the computation of future predictions.
There are classic examples of graphical models that condition predictions both on inputs at the current timestep and outputs at the previous~\cite{MEMM}.
Recent work on explicit checklists in neural settings, such as by~\cite{checklist} found success using a structured agenda that tracked ingredient usage in conditional cooking recipe generation.
Our checklists are however more transient; outputs can be removed from the checklist if enough time has passed.
We also track not just the presence of outputs, but information about how they were used.

There is also much work on the use of REINFORCE~\cite{REINFORCE} for sequence generation tasks.
While~\cite{Ranzato2016SequenceLT} performed several tasks including summarization, translation, and image captioning, but required a critic model for stabilization,~\cite{Rennie2017SelfCriticalST} developed a self-critical method for captioning that did not require this additional network.
\cite{Paulus2018ADR} applied a similar technique to abstractive news summarization.
Reinforcement learning has also been applied in a musical setting~\cite{musicrl}, including the REINFORCE algorithm specifically~\cite{seqgan,organ}, but most of these have focused on music generation rather than downstream prediction of performance-oriented attributes such as fingerings.

\vspace{-1ex}
\section{Model}

We now describe the basic layout of the models we compare, and detail the specific variations between them.
First, we will formalize the basic model contract and explain the basic architecture components that most of them share.

At a high level, each model takes in a MIDI-derived representation of either the left or right hand of a complete piano song, and outputs a predicted finger for each note.
More formally, we are given an input sequence of notes $\mathbf{x} = \{x_1, ..., x_N\}$, where each $x_i=(p_i,t^{-}_i,t^{+}_i)$ is a tuple consisting of a pitch $p_i \in \{1, ..., 88\}$ as well as a corresponding onset time $t^{-}_i \in \mathbb{R}_{\geq 0}$ and offset time $t^{+}_i > t^{-}_i$.
From $\mathbf{x}$ we must predict a corresponding sequence $\mathbf{y} = \{y_1, ..., y_N\}$ where each $y_i \in \{1,2,3,4,5\}$ is an index representing the finger used to play $x_i$ from thumb to pinky respectively.

To this end, each predictive model defines a distribution $p(\mathbf{y} | \mathbf{x} ; \theta)$, generally parameterized by an LSTM variant.
Under our non-checklist simple Bi-LSTM baseline, the individual $y_i$ are conditionally independent given $\mathbf{x}$:
$p(\mathbf{y} | \mathbf{x} ; \theta) = \prod^{N}_{i=1} p(y_i | \mathbf{x} ; \theta)$.
However for our autoregressive models, the likelihood factorizes according to chain rule:
$p(\mathbf{y} | \mathbf{x} ; \theta) = \prod^{N}_{i=1} p(y_i | \mathbf{y_{<i}} , \mathbf{x} ; \theta)$

As shown in Figure~\ref{fig:model}, the notes are first embedded, and then passed to a Bi-LSTM which outputs a contextualized representation.
This contextualized representation is optionally concatenated with a $d$ dimensional vector representation of the checklist $\mathbf{c} \in \mathbb{R}^{N \times d}$ before being passed to an MLP which outputs a softmax distribution over finger indices.
We can think of the checklist as a function of the past and current notes as well as the past finger predictions $c_i(\mathbf{x}_{\leq i},\mathbf{\hat{y}}_{<i})$.
The model is trained to optimize the cross entropy loss between this distribution and the true labels.

\begin{figure}
    \centering
    \includegraphics[trim= 20 210 25 210, clip, width=0.95\linewidth]{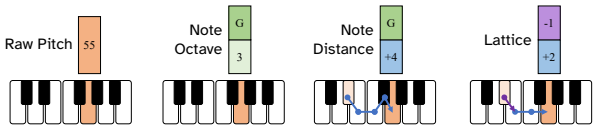}
    \vspace{-2ex}
    \caption{Examples of the input representations we compare. Note that the labels on the vectors are \textit{indices} corresponding to sub-embeddings shared across notes.}
    \label{fig:inputs}
    \vspace{-2ex}
\end{figure}

\subsection{Input Representations}

We now discuss four different strategies to encode the song into an input sequence of vector embeddings (illustrated in Figure~\ref{fig:inputs}).
These each expose different information to the model which can significantly affect performance.

\textbf{Raw Pitch:}
This approach simply represents each note by its pitch $p_i$, assigning a learnable embedding to each value.
This was done by most previous neural implementations~\cite{PIG,goldberg}, and requires learning $88$ vector embeddings.

\textbf{Note $\oplus$ Octave:}
Given the fact that a piano keyboard's layout is identical across octaves, we might believe that more important than the exact key being played is the named note itself (e.g. knowing that the next note is a \texttt{G\#} might be more useful in predicting which finger to use than knowing that it is the MIDI pitch $68$).
To this end we also evaluate the use of an input representation that consists of an embedding corresponding to the named note concatenated with an embedding corresponding to the octave.

\textbf{Note $\oplus$ Relative Distance:}
If finger choice is primarily a function not of the note's pitch but of its location on the keyboard relative to the other notes in the song, then directly exposing that information saves the model from having to infer it.
For this representation, we concatenate an embedding for the named note with one corresponding to the change in pitch $p_i - p_{i-1}$ (see Figure~\ref{fig:inputs}).
Since large distances between notes require physically lifting the hand from the keys, an act which eliminates any sequential constraints, we cap step sizes at $15$ semitones.
\cite{pitchdiff} used a similar approach without the named note embedding.

\textbf{Lattice:}
We finally evaluate a ``lattice'' representation based on the one introduced by~\cite{PIG} (note that this was only used in that work for HMMs and not neural models).
As shown in Figure~\ref{fig:inputs}, we can think of this as a two-dimensional encoding of relative position, where the first dimension corresponds to the number of white keys between notes (i.e. the \textit{horizontal} distance), and the second dimension indicates if we have moved from a white key to a black key or vice versa (i.e. the \textit{vertical} distance).
These dimensions are each embedded with corresponding vector embeddings, with a horizontal cutoff of $9$ steps.
This is the default representation where not otherwise specified.

\subsection{Checklist Formulation}

To encourage the model to produce fluent predictions, rather than simply choosing the most likely finger at each timestep independently, we also consider a series of extensions that feed recent predictions into the MLP alongside the contextualized embeddings produced by the LSTM, in the form of a vector embedding $c_i(\mathbf{x}_{\leq i},\mathbf{\hat{y}}_{<i}) \in \mathbb{R}^d$.

\textbf{Autoregressive Tagger:}
We start with a baseline autoregressive variant of our simple Bi-LSTM model in which the forward half of the Bi-LSTM acts as an encoder over prior predictions as well as notes.
In this setup, we do away with the Bi-LSTM over notes $\mathbf{x}$, and instead replace it with a forward LSTM $f(\mathbf{x},\hat{\mathbf{y}})$, and a backward LSTM $b(\mathbf{x})$.
At each timestep $i$, the MLP is fed the concatenated output of both LSTMs $f_i(\mathbf{x}_{<i}, \mathbf{y}_{<i}) \oplus b_i(\mathbf{x}_{\geq i})$.

\textbf{Prev Finger Embedding:}
As a precursor to a full checklist, we also experiment with a neural variant of a Maximum Entropy Markov Model (MEMM)~\cite{MEMM}.
Under this setup, we map the predicted label of the previous timestep $\hat{y}_{i-1}$ to a corresponding vector embedding, and concatenate this with the contextualized embedding of $x_i$ produced by the Bi-LSTM, before passing the result to the MLP.
This approach only takes into account the most recently used finger and ignores timing, but also requires the fewest additional model parameters.

\textbf{Binary Checklist:}
This checklist embedding is a $5$ dimensional vector, where each dimension corresponds to a finger.
The $i$th dimension is set to $1$ if that finger has recently been predicted for a note of a higher pitch than the current, a $-1$ if it has recently been predicted for a lower note, or a $0$ otherwise.
We empirically find that ``recent'' is best defined as within $100$ milliseconds.

\textbf{Distance Checklist:}
This representation is a concatenation of $5$ vector embeddings, one for each finger.
These finger embeddings use the lattice input representation strategy, where the first half corresponds to the horizontal distance between the note that that finger was recently predicted for and the current one, and the second half corresponding to the vertical distance.
If a finger has not recently been used, its section in the checklist vector will be filled by a learnable ``dummy'' embedding.

\begin{figure}
    \centering
    \includegraphics[trim= 15 210 15 88, clip, width=0.9\linewidth]{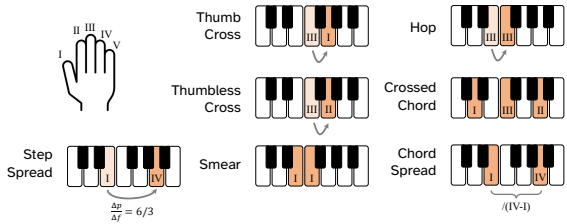}
    \vspace{-2ex}
    \caption{Fingering patterns tracked by our fluency metrics. Arrows and shading indicate sequential notes.}
    \label{fig:metrics}
    \vspace{-2ex}
\end{figure}

\section{Learning and Evaluation}

Next we will detail the additional metrics that we introduce, and show how by using reinforcement learning we can train our model to optimize some of them directly.

\subsection{Fluency Metrics}
\label{sec:metrics}

We now describe the metrics we use to measure model performance.
While research into pianists' fingering strategies shows that decisions are mostly motivated by minimizing ergonomically unfavorable patterns~\cite{SightReading,interview}, prior work has primarily evaluated on per-note labeling precision averaged over annotators, referred to as $M_{gen}$ (albeit with some different strategies for handling disagreements between annotators~\cite{PIG}).
However, by simply checking if each prediction agrees with any annotator's label for that note in isolation, we miss whether the model's predictions are coherent with one another.
For instance, a model may output two adjacent fingers which each agree with an annotator's label, but not the same one, creating a pattern that is harder to play than if it had solely aligned with one of them, despite these having the same $M_{gen}$ score.
Also, a model may output a sequence that agrees with none of the annotators but is at least playable, and yet receives the same score as a physically impossible sequence.
This is especially an issue for songs labeled by fewer annotators.

To address this, we expand the scope of our evaluation to include metrics that also measure how \textit{fluent} model predictions are within themselves.
To this end, we track statistics on the frequency of several types of output pattern that would increase the physical difficulty for a human performer, but may not be reflected in the raw labeling precision score.
This allows us to more holistically compare models in ways that are both attentive to sustained agreement with annotators, and also expose the specific ways in which their predictions are more or less playable.
Visual examples of these are provided in Figure~\ref{fig:metrics} for clarity.

\textbf{4-gram:}
Anchored 4-gram precision is the proportion of subsequences of four consecutive notes that were all predicted correctly with respect to a single annotator.
This metric measures output coherence, but is still directly tied to the gold labels.
Where $M_{gen}$ scores plateau as they approach the inter-annotator agreement of $71.4$, 4-gram precision has more headroom and better stratifies models.

\textbf{Thumb/Thumbless Crosses and Crossed Chords:}
Thumb crosses are where a note lower in pitch than the previous is played with a ``higher'' finger or vice versa, one of those two fingers being a thumb.
These shift the hand up or down the keyboard without fully lifting off of the keys.
We also track crosses where neither finger is a thumb, which is much less common due to their difficulty.
We refer to cases where two fingers are crossed and the notes overlap in time as crossed chords.
These are only performed under very specific circumstances, as most chords are not physically playable with crossed fingers.

\textbf{Hop:}
These are cases where one finger is used to play two different consecutive notes.
While not intuitively difficult, these do not allow the player to rest their hand in a fixed spot, and make distances harder to accurately judge, limiting the maximum tempo of a piece, similar to how hunt-and-peck typing is generally slower.

\textbf{Smear:}
In a smear, more than one note within a chord is played by a single finger.
These are generally utilized when two adjacent keys are both in the chord, but because of the placement of the other notes it is hard to use two separate fingers.
While these are extremely rare in our gold labels, naive baselines tend to produce them quite frequently.

\textbf{Step/Chord Spread:}
Step spread tracks how far the player must stretch their fingers apart while playing.
Specifically, it is the average number of semitones per finger separating any two adjacent but not overlapping notes.
For example, if a song contains an \texttt{E} followed by the \texttt{B} a fifth above it, the step spread would be $7$ for a model that predicts playing them with fingers $1$ and $2$ respectively, but only a $3.5$ for a model that predicts $1$ and $3$.
We measure similar cases where the notes do overlap in time as chord spread.
A large average chord spread is more challenging since it requires stretching the hand over a larger distance.

\begin{table*}
\centering
\resizebox{0.86\textwidth}{!}{
\begin{tabular}{rcccccccccc}
\toprule
Model & \thead{Thumb \\ Cross \texttt{(\#)}} & \thead{Thumbless \\ Cross \texttt{(\#)}} & \thead{Crossed \\ Chord \texttt{(\#)}} & \thead{Hop \\ \texttt{(\#)}} & \thead{Smear \\ \texttt{(\#)}} & \thead{Step Spread \\ \texttt{(s/f)}} & \thead{Chord Spread \\ \texttt{(s/f)}} & \thead{4-gram \\ \texttt{(Acc)}} & \thead{$M_{gen}$ \\ \texttt{(Acc)}}  \\
\midrule
Gold                              & 331 & 58 & 28 & 155 & 7 & 2.78 & 2.64 & 100 & 100 \\
\midrule
Bi-LSTM (MIDI)~\cite{PIG}$\dagger$& 228 & 84 & 47 & 485 & 148 & 2.63 & 2.56 & 66.7 & 27.8 \\
\midrule
Bi-LSTM (Note $\oplus$ Octave)    & \textbf{257} & 90 & \textbf{36} & 394 & 135 & \textbf{2.69} & 2.61 & 67.0 & 28.3 \\
Bi-LSTM (Note $\oplus$ Rel Dist)  & 218 & 87 & 52 & 390 & 101 & 2.65 & 2.59 & 69.1 & 32.2 \\
Bi-LSTM (Lattice)                 & 217 & \textbf{43} & 40 & \textbf{334} & \textbf{74} & 2.66 & \textbf{2.62} & \textbf{69.6} & \textbf{33.1} \\
\bottomrule
\end{tabular}
}
\caption{Results on PIG test set for Bi-LSTM model with different input representation schemes. The first row represents the value of each metric under the gold fingerings produced by human annotators. Values in bold are the closest to the gold labels' score for that metric. $\dagger$This model is based specifically on the neural approach used in~\cite{PIG}, but is our own implementation of it. It was similarly reimplemented by~\cite{goldberg} in their work. Units for most metrics are simple frequency counts, except for spreads which are in semitones per finger, and 4-gram and $M_{gen}$ which are accuracies.}
\label{tab:bilstm}
\vspace{-2ex}
\end{table*}

\subsection{Loss Functions}
\label{sec:reinforce}

Having defined our model as above, we can train it on the cross entropy loss between the predicted distribution $p(\mathbf{y} | \mathbf{x} ; \theta)$ and the gold labels by simply backpropagating to obtain parameter gradients and taking gradient descent updates on our training set.
Autoregressive models are teacher forced at train time (i.e. conditioned on gold labels from prior timesteps in lieu of the model's own previous predictions), and we decode with beam search at test time.

However, while training on cross entropy can yield strong performance on labeling precision, we find that this is not consistently correlated with other non-differentiable metrics which we may also wish to optimize with respect to.
We therefore also investigate using a supplemental loss function based on the REINFORCE algorithm~\cite{REINFORCE} which measures the frequency of undesirable fingering patterns in predicted outputs.
While REINFORCE is traditionally associated with environment navigation, it has seen success in sequence generation tasks as well~\cite{Ranzato2016SequenceLT,Rennie2017SelfCriticalST,Paulus2018ADR}.

Under this framework, the loss function is formulated as an expected reward, which we approximate by sampling a sequence $\tilde{\mathbf{y}}$ from our output distribution (using ancestral sampling) and weighting its reward by its likelihood.

\begin{equation*}
    L = \frac{(\bar{r} - r(\mathbf{x}, \tilde{\mathbf{y}}))}{N} \sum^{N}_{i=1} \log p(\tilde{y}_i | \mathbf{\tilde{y}}_{<i} , \mathbf{x} ; \theta)
\end{equation*}

Where $\bar{r}$ is the rolling average reward over the past $50$ examples, serving as our reward baseline.
In order to ensure stronger signal during training, we also calculate the REINFORCE loss over $10$ note chunks at a time, rather than over an entire song as this would make the attribution of reward to specific decisions less direct.
In our experiments, we set the reward to be a function of the number of hops and smears (see Section~\ref{sec:metrics}):

\begin{equation*}
\begin{split}
    r(\mathbf{x}, \mathbf{y}) = \exp( &- \#\{i : x_i \neq x_{i+1} , y_i = y_{i+1} \} \\
                                      &- \#\{ i : y_i = y_{i+1} , t^{+}_i > t^{-}_{i+1} \} )
\end{split}
\end{equation*}

Because training solely on the REINFORCE objective can lead the model towards degenerate solutions that trivially minimize undesirable patterns while compromising predictive accuracy, we take an approach similar to~\cite{Paulus2018ADR} and~\cite{Wu2016GooglesNM} where both loss functions are summed into a mixed training objective.
This provides a reasonable tradeoff between the predictive signal of cross entropy, and the pressure towards fluent output given by REINFORCE.
Mixed objective runs are warm-started on just the cross entropy loss to avoid degenerate solutions.

\vspace{-1ex}
\section{Experimental Setup}

\begin{table*}
\centering
\resizebox{0.82\textwidth}{!}{
\begin{tabular}{rcccccccccc}
\toprule
Model & \thead{Thumb \\ Cross \texttt{(\#)}} & \thead{Thumbless \\ Cross \texttt{(\#)}} & \thead{Crossed \\ Chord \texttt{(\#)}} & \thead{Hop \\ \texttt{(\#)}} & \thead{Smear \\ \texttt{(\#)}} & \thead{Step Spread \\ \texttt{(s/f)}} & \thead{Chord Spread \\ \texttt{(s/f)}} & \thead{$M_{gen}$ \\ \texttt{(Acc)}} & \thead{4-gram \\ \texttt{(Acc)}} \\
\midrule
Gold                              & 331 & 58 & 28 & 155 & 7 & 2.78 & 2.64 & 100 & 100 \\
\midrule
Bi-LSTM (MIDI)~\cite{PIG}$\dagger$               & 228 & 84 & 47 & 485 & 148 & 2.63 & 2.56 & 66.7 & 27.8 \\
HMM-3~\cite{PIG}$\ddagger$ & 220 & 32 & 31 & 196 & 84 & 2.84 & 2.72 & \textbf{70.2} & \textbf{39.5} \\
\midrule
Autoregressive Tagger             & 278 & 88 & 49 & \textbf{168} & 58 & 2.74 & 2.67 & 68.3 & 36.7 \\
(+REINFORCE)                      & 274 & 66 & 49 & 59 & 18 & 2.70 & 2.63 & 68.7 & 36.5 \\
\midrule
Prev Finger Embedding             & 271 & \textbf{56} & 40 & 127 & 30 & \textbf{2.78} & 2.67  & 68.1 & 35.5 \\
(+REINFORCE)                      & 283 & 53 & 54 & 53 & \textbf{12} & 2.81 & 2.76 & 68.4 & 36.5 \\
\midrule
Binary Checklist                  & 227 & 72 & \textbf{26} & 332 & 74 & 2.67 & 2.58 & 69.3 & 35.8 \\
(+REINFORCE)                      & 261 & 49 & \textbf{26} & 217 & 42 & 2.69 & \textbf{2.64} & 69.5 & 37.1 \\
\midrule
Distance Checklist                & 274 & 72 & 37 & 195 & 37 & 2.80 & 2.69 & 68.8 & 36.6 \\
(+REINFORCE)                      & \textbf{284} & 78 & 41 & 134 & 32 & 2.75 & 2.68 & 69.0 & 36.1 \\
\bottomrule
\end{tabular}
}
\caption{Results on PIG test set for various autoregressive setups, shown with and without using REINFORCE to minimize hops and smears (see Section~\ref{sec:reinforce}). $\ddagger$ A third order HMM implementation by~\cite{PIG} is also included -- while it ranks similarly as in prior work against LSTMs by $M_{gen}$, it falls behind on other metrics.}
\vspace{-2ex}
\label{tab:checklist}
\end{table*}

\subsection{Implementation Details}

We use $2$ layer LSTMs and $2$ layer MLPs with a hidden size of $1024$, and a dropout of $0.2$ in both of these networks.
We use $d$=$256$ for our input embedding size.
Beam search is performed to decode autoregressive models at test time using a beam size of $10$.
Models are trained using the Adam optimizer~\cite{adam} with a learning rate of $1$e$-4$.
Our code is implemented in PyTorch \texttt{1.8.1}~\cite{pytorch} and trains on an NVIDIA 2080ti GPU in roughly $12$ hours.

\subsection{Data}

We train and evaluate on the PIano fingerinG dataset (PIG)~\cite{PIG} which contains $150$ piano songs written by 24 composers.
Each song has up to $6$ fingerings produced by human pianists, yielding $309$ annotated songs in total.
Because of the imbalance in the dataset's original splits, we use alternate splits created by~\cite{goldberg} which increase the relative size of the train and validation sets.
As a baseline we compare to a reimplementation of the LSTM model of~\cite{PIG} (which was also reimplemented by~\cite{goldberg} in their experiments), albeit with the same architecture of our other systems.
We also show results from the third order HMM implementation of~\cite{PIG}, although the more similar neural models are our primary point of direct comparison.

We employ similar preprocessing as prior work.
Rather than modeling left and right hands separately, we reflect the pitches of all left hand parts and reverse the corresponding finger labels, thereby constructing a second ``right hand part'' which we treat as an independent song.
This prevents overfitting by halving the label space.
To handle chords, i.e. multiple notes with identical onsets, we simply arpeggiate them from lowest to highest pitch.
Otherwise, notes are ordered by onset.
We do however retain the original timing information for constructing accurate checklists.

\vspace{-1ex}
\section{Results}

\subsection{Input Representation Ablation}

We start by investigating the effects of input representations on a simple Bi-LSTM baseline model that outputs independent prediction probabilities at each timestep.
Table~\ref{tab:bilstm} shows our results across all metrics.
Overall the lattice representation does best, followed by note $\oplus$ octave.
This matches our hypothesis that learning a separate embedding for each note on the keyboard leads to overparameterization, and that modular representations are more effective.
We also see that relative distance based embeddings lead to fewer hops and smears specifically, which makes sense given that it allows the model to more easily observe when notes are or are not repeated, and therefore whether repeating the same finger would make sense.

\vspace{-1ex}
\subsection{Checklist Models}

We see in Table~\ref{tab:checklist} that the checklist models tend to do best overall, with Binary Checklist using REINFORCE getting the highest 4-gram score among LSTMs.
All the autoregressive systems significantly outperform the Bi-LSTM baselines, but with different tradeoffs in terms of the output patterns we measure.
REINFORCE minimizes the number of hops and smears compared to just cross entropy, and in doing so can sometimes boost other metrics as well.

Note that $M_{gen}$ fails to fully reflect what are at times substantial differences in model output made apparent from the other metrics.
For instance, the Prev Finger Embedding produces fewer hops, smears, and crossed chords than the Autoregressive tagger, likely indicating that it would be far easier for a human to perform despite having worse $M_{gen}$.
This is especially apparent when comparing the checklist models to the Bi-LSTM baselines.
Also while the HMM does strongly on $M_{gen}$ and 4-gram, we only see its deficits when we look at other metrics which reveal subtle issues such as an especially high finger spread and number of smears, and low utilization of thumb crosses.

\vspace{-1ex}
\subsection{Label Confusion}

\begin{wrapfigure}{r}{0.5\linewidth}
    \vspace{-1cm}
    \centering
    \includegraphics[trim= 220 180 220 170, clip, width=\linewidth]{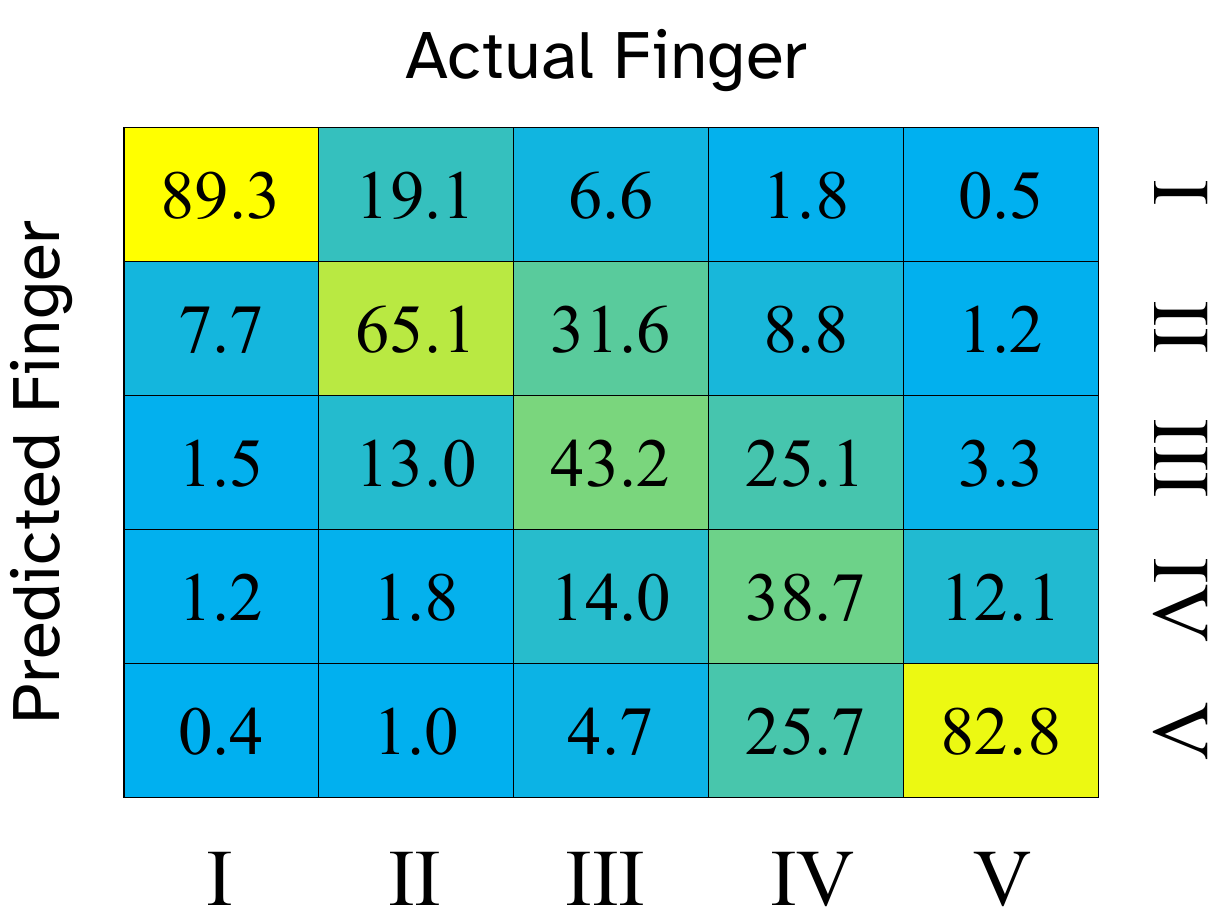}
    \vspace{-0.7cm}
    \label{fig:confusion}
\end{wrapfigure}

This figure shows the confusion matrix for predictions by the Binary Checklist model with REINFORCE.
The distribution of misclassifications is fairly different for each finger; adjacent fingers are more likely to be confused than ones that are further apart, reflecting a degree of interchangeability in how pianists use them.
We also see that thumb and pinky predictions are most accurate, perhaps because being at the ends of the hand makes them less versatile.

\subsection{Human Evaluation and Conclusion}

\begin{table}[H]
\resizebox{\linewidth}{!}{
\centering
\vspace{-1ex}
\begin{tabular}{rccc}
\toprule
Model & \thead{Physical \\ Comfort} & \thead{Mechanical \\ Efficiency} & \thead{Ease of \\ Learning} \\
\midrule
Bi-LSTM (MIDI)~\cite{PIG}       & 1.6 & 1.4 & 1.6  \\
Binary Checklist (+REINFORCE)    & \textbf{2.8} & \textbf{2.5} & \textbf{2.9}  \\
\bottomrule
\end{tabular}
}
\vspace{-2ex}
\caption{Human evaluation ratings across various criteria.}
\label{tab:human}
\end{table}

Since the ultimate success criteria of our task is to produce fingerings that are well-suited to human performance, we also present results from a small scale round of human evaluation.
Specifically, we recruited a college professor of piano with a doctorate in piano performance (who was not involved with the creation of the original dataset) to assess the outputs of two models---the Bi-LSTM (MIDI)~\cite{PIG} and Binary Checklist with REINFORCE---according to three main criteria: physical comfort, mechanical efficiency, and ease of learnability.
The pianist was asked to score the models' outputs with respect to each of these categories on a scale of 1-3 for all $10$ songs in the validation set (composed by Brahms, Mendelssohn, and Rachmaninoff).

We show the models' average ratings per category in Table~\ref{tab:human}.
The strong contrast suggests that the checklist model is not just predicting gold label fingers more frequently (shown by $M_{gen}$) but also that it is producing more locally coherent outputs in a way that substantially improves performability.
Furthermore, the correlation of human judgements with our proposed metrics indicates that they are meaningful measurements of output quality.

The pianist also provided qualitative observations which we summarize.
He noted that both systems performed best on simpler, repetitive passages such as those found in Mendelssohn.
Our model also reportedly produced fewer difficult stretches, which we expect as it is directly aware of interval distances in its input representation.
Another observed issue was that the Bi-LSTM's fingerings did not consider tempo, often containing sections that would be impossible to play fast enough, especially for Rachmaninoff pieces.
We suspect our model's advantage is from the checklist's dependence on absolute timing, not just the ordering of notes.
The pianist also reported that our model more frequently used the same fingering for repeated notes and chords, whereas the Bi-LSTM would often change fingerings without reason, which can also likely be explained by our conditioning on recent predictions instead of decoding them independently.
Quantitatively, we found that REINFORCE was able to reduce the number of hops and smears, which was confirmed by the pianist to have a significant impact on playability, emphasizing the importance of integrating these into the training loop instead of purely relying on a traditional likelihood loss.

Finally, he noted that the models particularly struggled on successive chords (as opposed to stepwise passages).
Chord transitions require fingerings that are not just comfortable in isolation but also connect together.
This requires the model to reason about \textit{voices} within the same hand, something which may require moving beyond a framework that treats songs as linear sequences of individual notes, and is more sensitive to polyphony.

\bibliography{ISMIRtemplate}

\end{document}